\documentclass[10pt,twocolumn,letterpaper]{article}

\usepackage{cvpr}              
\usepackage{multirow}
\usepackage{amssymb}

\DeclareMathOperator*{\argmin}{arg\,min}

\definecolor{cvprblue}{rgb}{0.21,0.49,0.74}
\usepackage[pagebackref,breaklinks,colorlinks,allcolors=cvprblue]{hyperref}

\def\paperID{} 
\def\confName{CVPR}
\def\confYear{2026}

\title{Prospective Dynamic 3D MRI Reconstruction via Latent-Space Motion Tracking from Single Measurement}

\author{Lixuan Chen \quad  Zhongnan Liu \quad  Jesse Hamilton \quad  James M. Balter  \quad 
Jeong Joon Park\footnotemark[2] \quad  Liyue Shen\footnotemark[2]\\
University of Michigan\\
}

\begin{document}
\maketitle

\footnotetext[2]{Co-corresponding senior authors who supervise this work together.}
\begin{abstract}
Prospective reconstruction is crucial in many clinical applications such as MRI-guided radiotherapy, which demands accurate image reconstruction and fast motion estimation from currently acquired measurements.
However, prospective reconstruction remains challenging due to ultra-sparse sampling and stringent latency requirements.
In this work, we propose PDMR, a \textbf{P}rospective \textbf{D}ynamic 3D \textbf{M}RI \textbf{R}econstruction framework with latent-space motion tracking.
Our core idea is to learn an efficient and generalizable latent manifold of motion fields offline, enabling rapid online adaptation for prospective reconstruction.
Specifically, we parameterize the deformation vector fields (DVFs) on a low-dimensional manifold, effectively reducing the search space for fast online adaptation, and employ a tri-plane representation to achieve geometry-aware and memory-efficient encoding of 3D motion.
Experiments on both XCAT digital phantoms and in-house abdominal MRI datasets demonstrate that PDMR achieves high-fidelity and temporally consistent reconstruction across multiple prospective scenarios (Immediate and After-2min), outperforming state-of-the-art 
retrospective and online 
methods. Our results suggest a promising pathway toward ultra-fast, motion-aware prospective MRI reconstruction in clinical practice.

\end{abstract}    
\section{Introduction}
\label{sec:intro}

\begin{figure}[t]
  \centering
   \includegraphics[width=\linewidth]{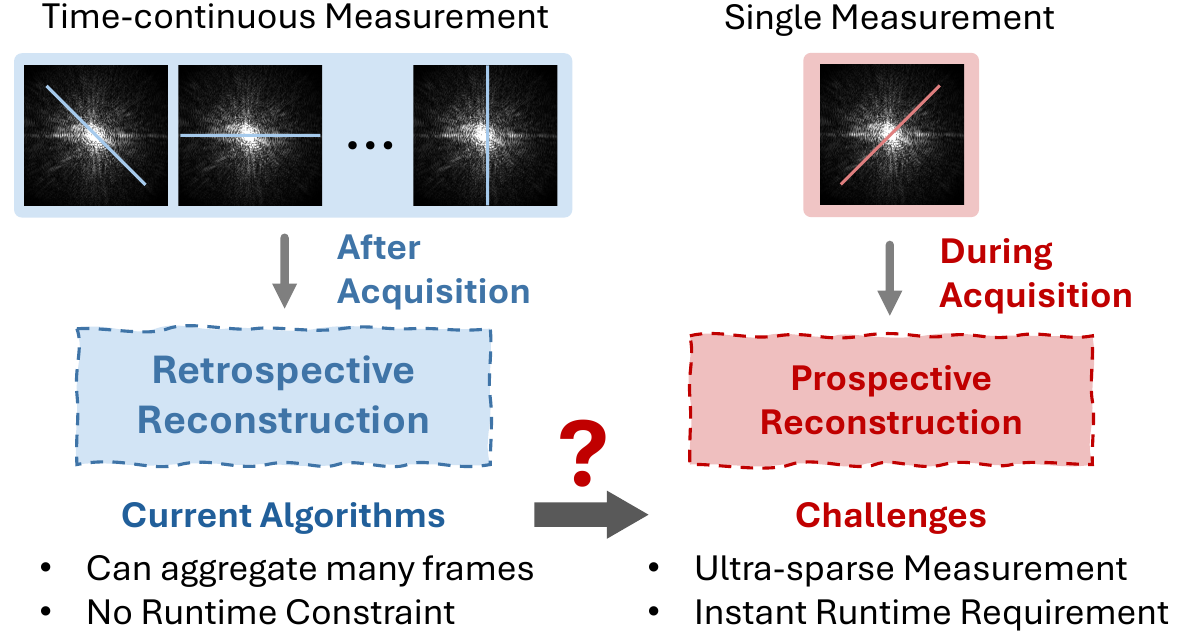}

   \caption{Retrospective methods face challenges in prospective reconstruction. 
   }
   \label{fig:teaser}
\end{figure}

In radiotherapy or interventional procedures, real-time visualization of 3D patient anatomy and accurate motion estimation are essential for enhancing treatment precision, as they directly guide therapeutic decisions~\cite{RT-MRI}.
Motivated by this need, our goal is to reconstruct images directly from instantaneously sampled ultra-sparse measurements acquired within the current latency window, known as \textit{prospective reconstruction}.

In recent years, MR-guided treatment, where magnetic resonance imaging (MRI) is used to guide radiotherapy procedure of cancer patients to mitigate the motion effects such as respiration, has increasingly been recognized as an advanced clinical option~\cite{keall2022integrated}.
Although MRI offers unparalleled soft-tissue contrast and non-invasive advantages, 
its inherently long acquisition time fundamentally limits the number of measurements that can be acquired within the allowed latency window for motion tracking.
The resulting ultra-sparse sample makes prospective reconstruction problem highly ill-posed.
Consequently, most previous works~\cite{GRASP, yoo2021time, chen2025single} focus on \textit{retrospective reconstruction}, where dynamic images are reconstructed from time-continuous measurements after acquisition by aggregating all time frames together for a joint dynamic reconstruction, leveraging the spatio-temporal redundancy.

However, existing retrospective approaches cannot satisfy the demands of ultra-fast, low-latency prospective reconstruction, due to the significant challenges of ultra-sparse measurement and instant runtime requirement in the setting of prospective reconstruction, as shown in Fig.~\ref{fig:teaser}.
On the other hand, the retrospective reconstruction has learned informative spatial-temporal representations offline for the specific patient, which naturally offers an important prior for prospective reconstruction of the same patient.
Therefore, a key question arises: \textit{how can we leverage the rich priors learned from retrospective reconstruction to enable low-latency prospective reconstruction from ultra-sparse measurement (i.e., a single-shot $k$-space measurement)?} 
This overarching question can be further divided into two sub-problems: (a) How can we learn a good retrospective prior?
(b) How can we rapidly adapt this prior for low-latency prospective reconstruction?

\paragraph{How to learn a good retrospective prior?}
Although retrospective reconstruction can exploit the full spatio-temporal redundancy across all time frames, learning a powerful prior from undersampled dynamic measurements still remains a very challenging problem.
One promising direction is motion-compensated (MoCo) representation~\cite{pan2022learning,pan2024motion,morales201implementation, tian2026unsupervised}, which explicitly decompose the dynamic sequence into a time-dependent deformation vector field (DVF) and a static template image.
This decomposition allows the entire image sequence to share the same spatial structure and effectively exploit the spatio-temporal redundancy in the acquired measurements, leading to improved reconstruction quality.
Clinically, patients often undergo multiple scans across treatment fractions.
The MoCo decomposition naturally allows the template image to be obtained from a pre-scan or a previous fraction, providing high-quality patient-specific structural information as a prior, further alleviating the ill-posedness.

\paragraph{How can we rapidly adapt this prior for low-latency prospective reconstruction?}
With the MoCo decomposition, prospective reconstruction reduces to learning a patient-specific motion representation that can be quickly adapted to a new motion status at each new time point.
Recent methods attempt to address this.
MR-MOTUS~\cite{MR-MOTUS} and DREME-MR~\cite{shao2025dynamic} model deformation vector fields (DVFs) as a {\em linear} combination of spatial and temporal bases, enabling compact motion representations from undersampled measurements. During prospective reconstruction, they reuse retrospectively learned spatial bases and update only the temporal coefficients, allowing efficient online adaptation. However, because linear bases cannot capture the complex, nonlinear anatomical motion seen in practice, their representational limits reduce accuracy and robustness under ultra-sparse sampling.
Prior-INR~\cite{liu2024volumetric} takes a different strategy by defining a discrete prior space spanning representative respiratory states. Online reconstruction searches this space to find the prior that best matches the incoming measurement. While effective under simplified motion patterns, this handcrafted and discrete manifold does not generalize well and fails to reflect the continuous nature of physiological motion, limiting its clinical applicability.

In this work, we propose \textbf{PDMR}, which is, to the best of our knowledge, the \textit{first} \textbf{P}rospective \textbf{D}ynamic \textbf{M}RI \textbf{R}econstruction framework that leverages a non-linear manifold-based deformation representation.
To enable effective adaptation from a single measurement during prospective reconstruction, we first retrospectively learn a compact and temporally generalizable latent space that can capture the underlying dynamics of deformation fields.
Specifically, we incorporate a geometry-aware mapping network based on the tri-plane representations~\cite{chan2022efficient} to learn a nonlinear mapping from the latent manifold to full 3D DVFs. 
The tri-plane representation provides a high-resolution and structurally coherent feature embedding, allowing the model to preserve both global anatomy and fine local deformation details. This results in a strong retrospective prior that makes fast and stable DVF adaptation feasible during prospective reconstruction.

Benefiting from the compact manifold and the generalizable mapping network, PDMR can efficiently and accurately estimate the current DVF from instantaneous measurements (\ie, a single spoke). 
During prospective reconstruction, PDMR only needs to optimize a low-dimensional latent vector on the learned manifold, requiring just a few iterations to recover the current motion state and produce high-quality images.
Experiments on both XCAT digital phantoms~\cite{XCAT} and in-house abdominal MRI datasets demonstrate that PDMR achieves high-fidelity reconstruction and accurate motion tracking across multiple prospective scenarios, confirming the effectiveness and generalizability of manifold-based DVF modeling.

\par To summarize, our key contributions are as follows:
\begin{itemize}
    \item We introduce PDMR, the first framework that leverages a retrospectively learned, patient-specific motion manifold that can rapidly adapt to the latency window, enabling high-quality prospective reconstruction from instantaneous measurements.
    \item We propose a new geometry-aware manifold mapping network tailored to estimate a fine-grained time-varying deformation field from ultra-sparse measurements.
    \item We evaluate PDMR extensively, showing superior reconstruction quality and robustness to diverse motions across multiple prospective MR scenarios.  
\end{itemize}
\section{Related Work}
\label{sec:relat}

\subsection{Manifold Learning in Medical Imaging}
Manifold learning has been widely used to capture the low-dimensional structure underlying high-dimensional medical images. Classical nonlinear methods such as LLE \cite{roweis2000nonlinear}, Laplacian Eigenmaps \cite{belkin2003laplacian}, and diffusion maps \cite{coifman2006diffusion} have been applied to analyze anatomical shape variability and deformation field \cite{MR-MOTUS,shao2025dynamic}. With deep learning, generative models, including VAEs \cite{kingma2013vae} and latent-variable motion models for cardiac and respiratory dynamics \cite{beetz2022interpretable}, learn smooth latent spaces that capture physiological states \cite{brosch2013manifold}. These representations have enabled tasks such sparse-view reconstruction, or motion modeling from limited observations.

More recently, implicit neural representations (INRs) have emerged as continuous neural manifolds for encoding anatomy and motion \cite{park2019deepsdf,sitzmann2020implicit,mildenhall2021nerf,liu2024volumetric,shen2022nerp}. INR-based models~\cite{feng2025spatiotemporal} have been used for volumetric MRI reconstruction and deformation modeling, but most operate retrospectively or depend on fixed priors that cannot rapidly adapt to new measurements.

Overall, while manifold learning effectively represents complex motion in low-dimensional spaces, existing approaches are not designed for prospective, ultra-sparse, or online MRI settings, motivating the proposed method.

\subsection{Prospective MRI Reconstruction}
Prospective MRI reconstruction is essential for real-time visualization during interventional and image-guided procedures. By providing clinicians with immediate imaging feedback, for example in MRI-guided radiotherapy, prospective methods can improve targeting accuracy, enhance tumor treatment efficacy, and reduce radiation exposure to surrounding healthy tissues.

Several recent works~\cite{liu2024volumetric, MR-MOTUS, shao2025dynamic} have explored patient-specific modeling to address this challenge. Liu \textit{et al.}~\cite{liu2024volumetric} proposed a prior-augmented INR that learns a continuous volumetric representation from only two fully sampled 3D MRIs, enabling sparse 2D cine slices to be lifted into full 3D volumes. Although effective under stable breathing conditions, the method heavily depends on the two static priors, making it sensitive to changes in respiratory patterns. Furthermore, the smoothness bias inherent to the INR formulation limits its ability to represent nonlinear or discontinuous physiological motion, leading to unreliable extrapolation when the motion deviates from the training distribution.

Huttinga \textit{et al.}~\cite{MR-MOTUS} reconstruct respiratory motion directly from undersampled 3D $k$-space by enforcing a low-rank linear motion model. This enables sub-200 ms inference; however, the method assumes that all non-rigid 3D deformations can be expressed as a linear combination of a few fixed spatial bases. Such a linear assumption limits robustness in the presence of nonlinear organ motion or sliding interfaces, where the true deformation cannot be accurately represented by a low-rank linear subspace.

Despite these advances, prospective MRI reconstruction remains largely underexplored. Existing approaches either rely on restrictive priors, impose strong assumptions on motion linearity, or fail to maintain accuracy when motion deviates from the training distribution. Achieving ultra-fast and anatomically reliable volumetric reconstruction therefore remains a critical unmet need in the field.

\begin{figure*}[t]
  \centering
   \includegraphics[width=\linewidth]{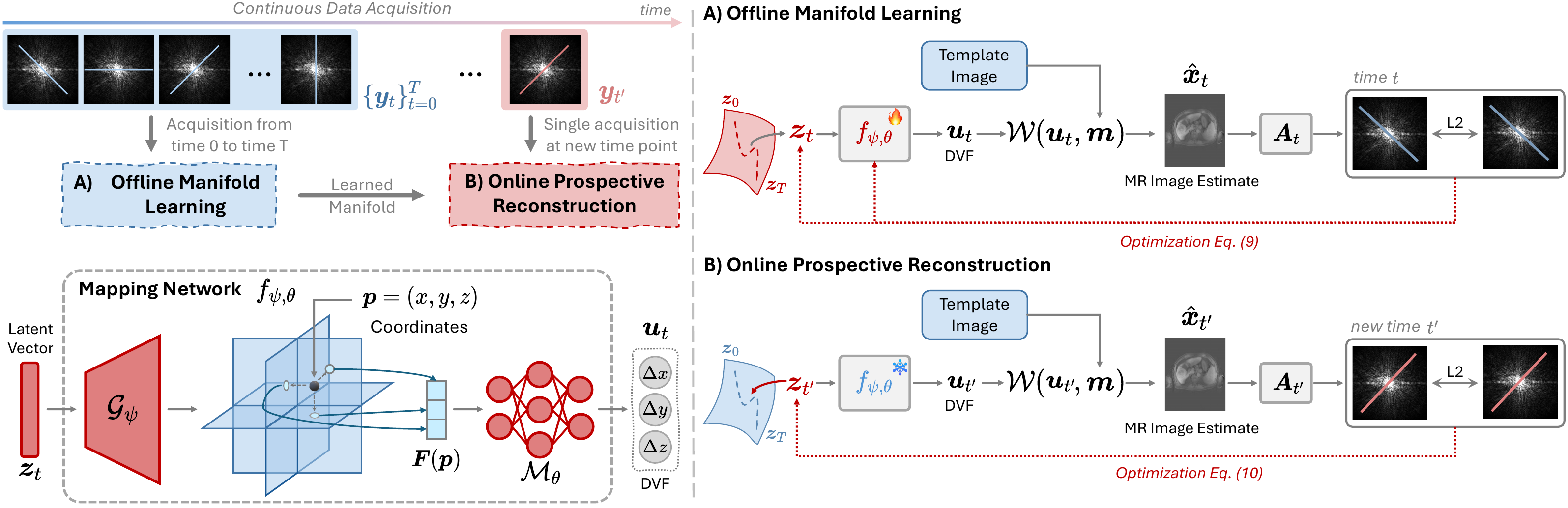}
   \caption{Overview of proposed PDMR.
\textbf{A.} PDMR performs offline manifold learning, where the patient-specific motion manifold and DVF mapping network $\boldsymbol{f}_{\psi, \theta}$ are learned from time-continuous sparse measurements $\{\boldsymbol{y}\}_{t=0}^T$ in a retrospective manner.
\textbf{B.} During online prospective reconstruction, given a single instantaneous measurement $\boldsymbol{y}_{t'}$, PDMR rapidly adapts by optimizing only the latent vector while keeping the learned mapping network fixed. Meanwhile, PDMR employs a geometry-aware tri-plane mapping network, enabling the latent vector to be mapped to fine-detailed DVFs efficiently and supporting fast adaptation.
}
   \label{fig:pipeline}
\end{figure*}

\section{Preliminaries}
\label{sec:method}

\subsection{Forward Model of Dynamic MRI}
The forward acquisition of dynamic MRI is formulated as: 
\begin{equation}
\boldsymbol{y}_{t}=\boldsymbol{P}_{t}\mathcal{T}\boldsymbol{x}_t+\boldsymbol{n}_t,
\label{eq:forward}
\end{equation}
where $\boldsymbol{x}_t \in \mathbb{C}^m$ is the image at timestamp $t$, $\mathcal{T}$ is the Fourier operator, $\boldsymbol{P}_{t}$ is the time-varying sampling pattern, and $\boldsymbol{y}_t \in \mathbb{C}^{n}$ is the acquired $k$-space (frequency domain of the image) measurements~\cite{model_based_mri}.  
To achieve high temporal resolution, the acquisition is typically highly undersampled in the spatial domain (\textit{i.e.}, $n \ll m$).
The goal of \textbf{retrospective reconstruction} is to recover a high-quality dynamic sequence  $\{\boldsymbol{x}_t\}_{t=0}^{T}$ from these undersampled measurements.

\subsection{Motion-Compensated (MoCo) Reconstruction}
MoCo-based methods decompose each dynamic image $\boldsymbol{x}_t$ into a time-varying deformation field $\boldsymbol{u}_t$ and a static template image $\boldsymbol{m}$, formulated as:
\begin{equation}
\boldsymbol{x}_t = \mathcal{W}(\boldsymbol{m}, \boldsymbol{u}_t),
\end{equation}
where $\mathcal{W}(\cdot)$ denotes the image warping operator. 
The DVF specifies, for each voxel in $\boldsymbol{x}_t$, the displacement vector $(\Delta x, \Delta y, \Delta z)_t$ that maps it to the corresponding location in the template image space.
By leveraging this decomposition, MoCo-based methods transform dynamic reconstruction into estimating the deformation fields with a shared template image.
The use of a shared template enables these methods to effectively exploit spatial–temporal correlations~\cite{RR_overview}, allowing high-quality reconstruction even from highly undersampled $k$-space data.

\subsection{Motion-Tracking Guidance Prospective Reconstruction}
For prospective reconstruction, the MoCo decomposition is particularly attractive: the template image can be pre-reconstructed using pre-scan data, providing a patient-specific structural prior. 
Consequently, the prospective reconstruction problem can be reformulated as estimating the current deformation field from the instantaneous $k$-space measurement, rather than reconstructing a full 3D image from scratch.
The optimization process can be expressed as: 
\begin{equation}
    \boldsymbol{u}^*_{t'} = \underset{\boldsymbol{u}_{t'}}{ \argmin} \left \| \boldsymbol{y}_{t'} - \mathcal{W}(\boldsymbol{m}, \boldsymbol{u}_{t'})\right \|_2^2
\end{equation}
where $t'$ denotes the new timepoint and $\boldsymbol{y}_{t'}$ denotes the instantaneous $k$-space measurement.

\section{Proposed Method}
Our goal is to reconstruct high-quality images in a prospective MR acquisition from instantaneous measurements that reflect the object’s current motion state.
To achieve this, we propose \textbf{PDMR}, which introduces a manifold representation (\S~\ref{sec:manifold_representation}) to enable fast online motion adaptation.
PDMR retrospectively learns a generalizable motion manifold with an effective mapping function from pre-scan dynamic data (\S~\ref{sec:offline_learning}). 
Leveraging the learned manifold and mapping function, PDMR requires optimizing \textbf{only a low-dimensional latent code} to estimate a high-quality deformation field, enabling fast adaptation to new motion states  (\S~\ref{sec:online_reconstruction}).

\subsection{Manifold-based DVF Representation}
\label{sec:manifold_representation}
Deformation vector fields (DVFs) reflect object movements driven by various physiological signals. 
To model these complex non-linear mappings, we propose a manifold-based DVF representation that formulates DVF as a function of a low-dimensional latent vector: 
\begin{equation}
    \boldsymbol{f}:\boldsymbol{z}\in \mathbb{R}^r \mapsto \boldsymbol{u}\in \mathbb{R}^{m\times3},
\end{equation}
where $\boldsymbol{z}$ encodes the underlying motion state and $\boldsymbol{u}$ is the corresponding deformation field.
To capture fine-grained deformation patterns and ensure efficient computation, we approximate the function $\boldsymbol{f}$ using a geometry-aware mapping network.

\paragraph{Geometry-Aware Mapping Network.}
Directly mapping the latent code to a full 3D DVF would introduce substantial computational and unstable optimization.
Therefore, we adopt a geometry-aware mapping network $\boldsymbol{f}_{\psi, \theta}$ composed of a tri-plane-based generator $\mathcal{G}_{\psi}$ and decoder $\mathcal{M}_{\theta}$ to approximate the function $\boldsymbol{f}$ for accurate DVF estimation, inspired by~\cite{chan2022efficient}.
\par Specifically, the tri-plane-based generator $\mathcal{G}_{\psi}$ first maps the latent vector $\boldsymbol{z}$ into three orthogonal feature planes $\{\boldsymbol{F}_{xy}, \boldsymbol{F}_{xz}, \boldsymbol{F}_{yz}\}$. 
Given a spatial coordinate $\boldsymbol{p} = (x, y, z)$, its corresponding features are retrieved from the three planes via projection. 
Then, the geometry-aware feature at $\boldsymbol{p}$ is obtained by concatenating these three planar features, formulated as:
\begin{equation}
    \boldsymbol{F}(\boldsymbol{p}) = \boldsymbol{F}_{xy}(x,y) \oplus \boldsymbol{F}_{xz}(x,z) \oplus \boldsymbol{F}_{yz}(y,z), 
\end{equation}
where $\oplus$ denotes feature concatenation. 
Next, the decoder $\mathcal{M}_{\theta}$ takes the fused feature as input and predicts the corresponding displacement vector at that position, (\textit{i.e.}, $\Delta \boldsymbol{p} = \mathcal{M}_{\theta} (\boldsymbol{F}\left ({\boldsymbol{p}}) \right )$).
Finally, by querying all spatial coordinates in the imaging space $\Omega$, we obtain the estimated DVF $\boldsymbol{u}$ as:
\begin{equation}
    \boldsymbol{u} = \left\{ \boldsymbol{f}_{\psi, \theta} (\boldsymbol{z}, \boldsymbol{p})\right\}_{\boldsymbol{p}\in\Omega}.
\label{eq:dvf_estimate}
\end{equation}

\subsection{Offline Manifold Learning}
\label{sec:offline_learning}
Figure~\ref{fig:pipeline} (A) shows the overall offline training process. 
To learn a compact manifold that can represent diverse motions and enable fast adaptation during prospective reconstruction, we jointly optimize latent manifold and network using pre-collected time-continuous measurements $\boldsymbol{Y} = \{\boldsymbol{y}_t\}_{t=0}^{T}$ and a template image $\boldsymbol{m}$, \eg from the pre-treatment scanning for the specific patient in standard radiotherapy procedure.

\par Specifically, for each timestep $t$, we first sample the latent vector $\boldsymbol{z}_t$ from a Gaussian distribution, assumed prior distribution. 
The mapping network then takes the latent vector and imaging spatial coordinates as input to estimate the time-varying DVFs $\hat{\boldsymbol{u}}_t$ (Eq.~\ref{eq:dvf_estimate}). 
Given the estimated DVFs, the template image $\boldsymbol{m}$ is warped to generate the dynamic image:
\begin{equation}
    \hat{\boldsymbol{x}}_t\ =  \mathcal{W}(\hat{\boldsymbol{u}}_t, \boldsymbol{m}), 
\end{equation}
where $\mathcal{W}(\cdot)$ denotes warping operation. 
Intergating the forward model of dynamic MRI (Eq.~\ref{eq:forward}), we can get the $k$-space estimates as:
\begin{equation}
    \hat{\boldsymbol{Y}} = \{\boldsymbol{A}_t \hat{\boldsymbol{x}}_t \}_{t=0}^{T}, \ \text{with} \ \boldsymbol{A}_t\triangleq\boldsymbol{P}_t\mathcal{T},  
\end{equation}
Finally, under an auto-decoder formulation~\cite{park2019deepsdf}, both the latent codes and the mapping network are optimized by minimizing the measurement consistency loss together with a DVF regularization term: 
\begin{equation}
\begin{aligned}
 \boldsymbol{Z}^*, \psi^*, \theta^* & = \underset{\boldsymbol{Z},  \psi, \theta}{\argmin}  \| \hat{\boldsymbol{Y}} - \boldsymbol{Y} \|_2^2 + \lambda \mathcal{R}(\boldsymbol{U}), \\
& \text{where} \ \boldsymbol{Z}=\{\boldsymbol{z}_t\}_{t=0}^{T}, \boldsymbol{U}=\{\boldsymbol{u}_t\}_{t=0}^{T}, 
\end{aligned}
\end{equation}
where $\mathcal{R}(\cdot)$ denotes the DVF regularization term that enforces temporal smoothness, and $\lambda$ is the weighting hyperparameter.

\subsection{Online Prospective Reconstruction}
\label{sec:online_reconstruction}

After completing the offline manifold learning stage, we obtain a compact manifold and a well-trained mapping network to capture patient-specific motion representations through retrospective training. 
Given a random sample from this manifold, the mapping network is able to generate fine-grained spatio-temporal deformation patterns. 
Therefore, in the online prospective reconstruction, we freeze the parameters $(\psi^*, \theta^*)$ of the mapping network and optimize only the latent vector $\boldsymbol{z}$ corresponding to the current time frame.

\begin{figure*}[t]
  \centering
   \includegraphics[width=\linewidth]{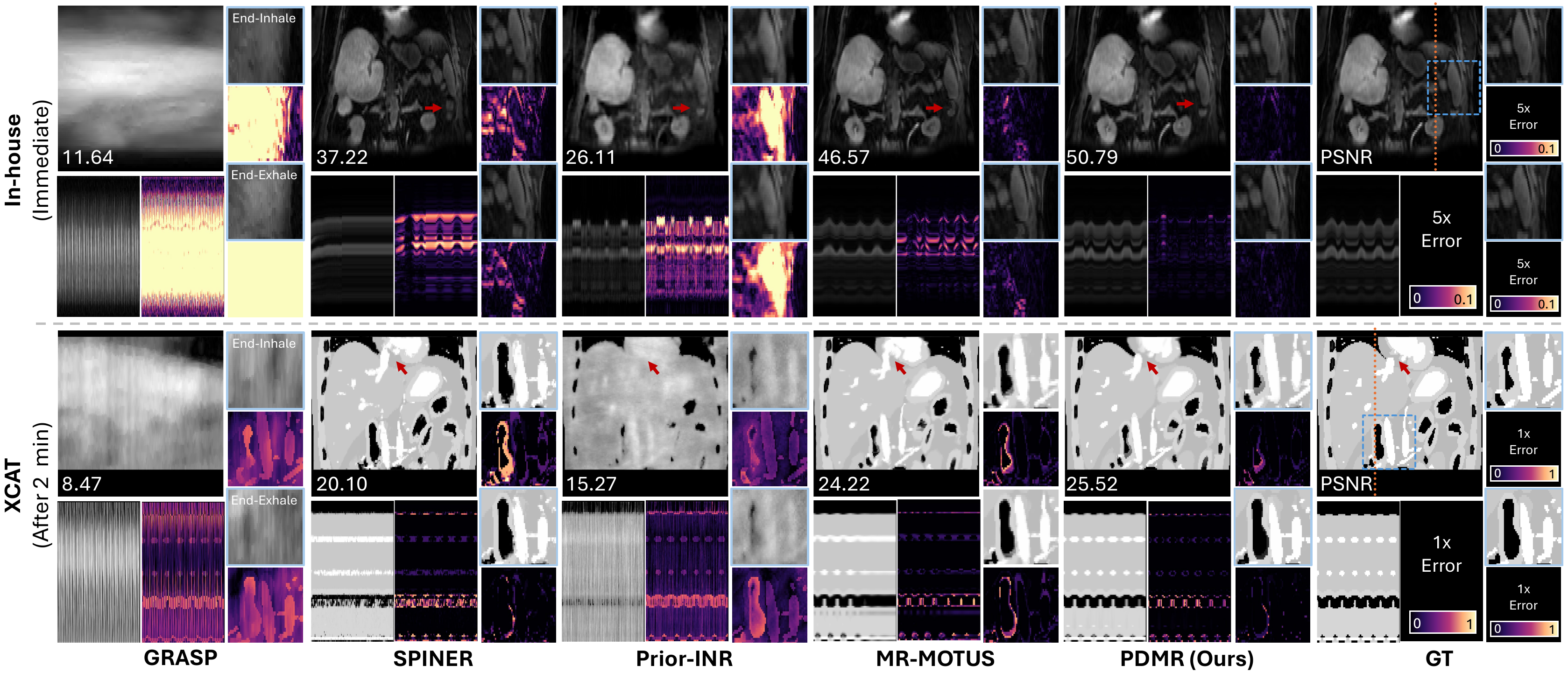}

   \caption{Qualitative comparisons of prospective reconstruction results on the in-house dataset (top row) and the XCAT dataset (bottom row). We display the reconstructed images, the over-time profile lines in the $z$–$t$ plane, and the corresponding error maps.
The selected $z$-axis location is marked by an orange dashed line, and zoom-in boxes highlight regions of interest at the end-inhale and end-exhale motion states for improved visualization of motion capture.
Red arrows indicate noticeable small-motion capture failures in the baselines. }
   \label{fig:pros}
\end{figure*}

Specifically, given an instantaneous measurement $\boldsymbol{y}_{t'}$, the latent vector $\boldsymbol{z}_{t'}$ is optimized within the learned manifold by minimizing the discrepancy between the estimated and acquired $k$-space data. 
The optimization process can be formulated as:
\begin{equation}
\begin{aligned}
       & \boldsymbol{z}_{t'} = \underset{\boldsymbol{z}}{\argmin}\left \| \boldsymbol{A}_{t'}\boldsymbol{x}_{t'} - \boldsymbol{y}_{t'}\right\|_2^2, \\ 
      & \text{with} 
    \ \boldsymbol{x}_{t'}= \mathcal{W}\left(\boldsymbol{m}, \boldsymbol{f}_{\psi^*,\theta^*}(\boldsymbol{z})\right).   
\end{aligned}
\label{eq:online}
\end{equation}
Once the optimal latent vector $\hat{\boldsymbol{z}}_{t'}$ is obtained, 
the corresponding DVF $\hat{\boldsymbol{u}}_{t'} = f_{\psi,\theta}(\hat{\boldsymbol{z}}_t)$ 
is used to warp the template image $\boldsymbol{m}$ to generate the current frame:
\begin{equation}
\hat{\boldsymbol{x}}_{t'} = \mathcal{W}(\boldsymbol{m}, \hat{\boldsymbol{u}}_{t'}).
\end{equation}
This procedure enables ultra-fast motion tracking and prospective reconstruction.
Since only a low-dimensional latent vector needs to be optimized per time frame, 
the inference process is highly efficient and can rapidly adapt to unseen motion states 
while preserving the physical plausibility encoded in the learned manifold.

\section{Experiments \& Results}
\label{sec:exp}

\begin{figure}[t]
  \centering
   \includegraphics[width=\linewidth]{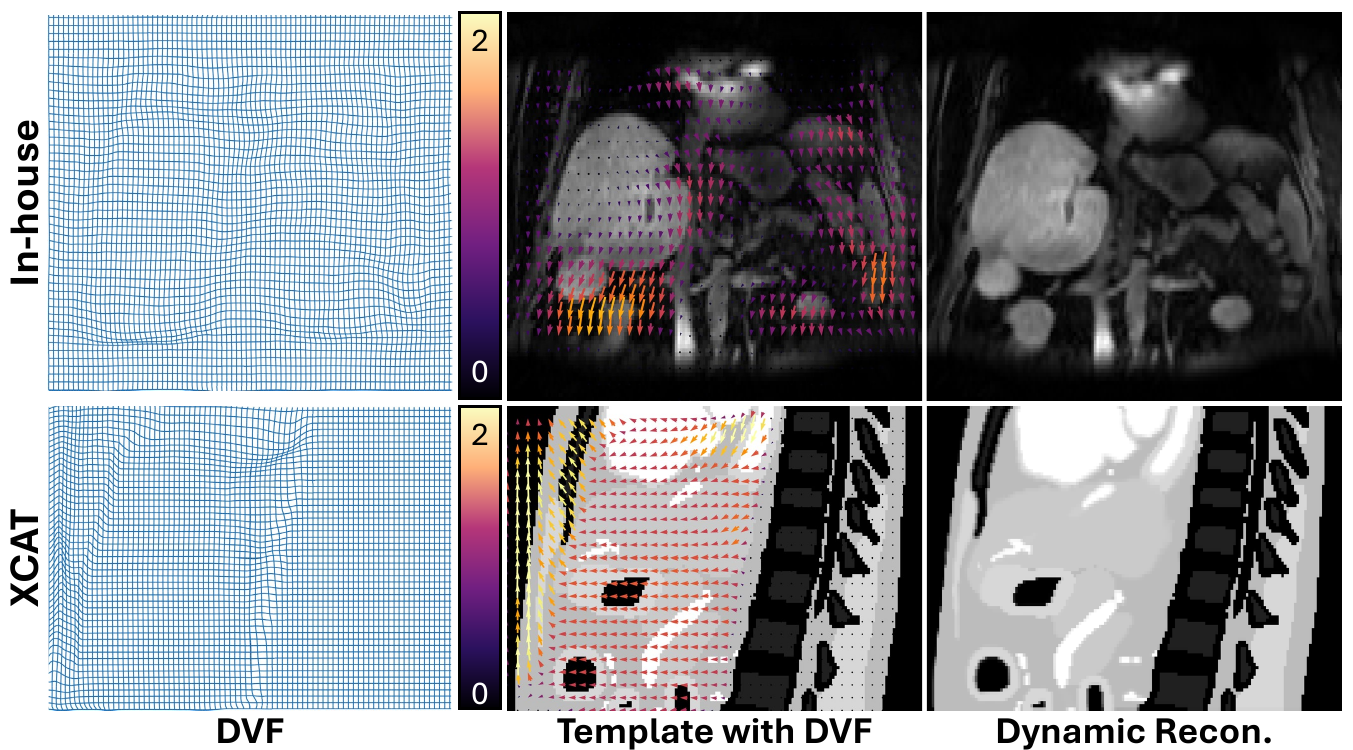}
   \caption{Visualization of the estimated DVFs during prospective reconstruction on the in-house dataset (coronal view) and the XCAT dataset (sagittal view).}
   \label{fig:dvf}
\end{figure}

\subsection{Datasets and Preprocessing}
\paragraph{XCAT Phantom Data.} 
We used digital XCAT phantoms~\cite{XCAT} with simulated respiratory motion in the abdominal region, where the parameter maps were assigned based on previous literature~\cite{chen2016mr} and the dynamic 3D data were simulated using a spoiled gradient echo sequence configured with the same acquisition parameters as our in-house protocol, serving as ground-truth reference data.
Each respiratory cycle lasted 3.97 seconds, and the dynamic sequence was simulated with a temporal resolution of 170 ms per frame.

\paragraph{In-house Data.}
In-house data were acquired from six subjects under an institutional review board (IRB)–approved protocol.
A 10-min DCE-MRI scan was performed using a work-inprogress golden-angle stack-of-stars radial trajectory, yielding 3500 spokes ($~$170 ms per spoke acquisition).
Using the hierarchical motion modeling framework~\cite{zhang2021hierarchical}, we extracted respiratory motion signals by analyzing the liver’s center-of-mass trajectory, and used these signals to bin the data and reconstruct 21 respiratory-resolved 3D volumes.
Each volume was deformably registered to the end-exhale reference to obtain DVFs for all 21 respiratory states.
Per-spoke DVFs were computed via linear interpolation between the bounding respiratory states based on its continuous respiratory position.
By applying these DVFs to the reference volume, we generated high–temporal-resolution dynamic 3D MRI datasets, which serve as ground truth.

\paragraph{Simulated Acquisition Process.}
For both the XCAT phantom and the six in-house dynamic 3D MRI datasets, we employed a golden-angle stack-of-stars sampling pattern, acquiring only one stack of spokes per time frame. 
Each spoke consisted of 448 readout samples with $k_z = 96$ partitions.
Spokes 0–150 were used for offline manifold learning, while prospective reconstruction was evaluated using spokes 150–300 and 1000–1150.
Since each spoke corresponds to approximately 170 ms, the latter interval reflects data acquired roughly 2 minutes after the initial acquisition.

\subsection{Baselines \& Metrics}
\paragraph{Baselines.} 
We compare PDMR with six representative dynamic MRI reconstruction approaches: 
\par 1) Analytical algorithms: 
\textbf{NUFFT}~\cite{NUFFT} performs non-Cartesian reconstruction using the non-uniform FFT; \textbf{GRASP}~\cite{GRASP} applies compressed-sensing constraints under golden-angle sampling to jointly reconstruct the dynamic sequence retrospectively; 
\par 2) Retrospective methods: \textbf{TDDIP}~\cite{yoo2021time} extends Deep Image Prior~\cite{DIP} to dynamic imaging by fitting a CNN generator to the full temporal measurements; \textbf{SPINER}~\cite{chen2025single} uses an implicit neural representation to model the dynamic volume as a continuous function of space and time; 
\par 3) Prospective methods: \textbf{Prior-INR}~\cite{liu2024volumetric} employs a hand-crafted discrete motion manifold; \textbf{MR-MOTUS}~\cite{MR-MOTUS} updates motion in a linear subspace using sequential $k$-space measurements.

\paragraph{Evaluation Metrics.} 
For the reconstructed dynamic MRI, we employ peak signal-to-noise ratio (PSNR) and structural similarity index (SSIM) as quantitative evaluation metrics.

\subsection{Implementation Details}

In PDMR, the latent code dimension is set to $r = 12$, and each tri-plane contains 32 feature channels. 
The mapping network is composed of a tri-plane encoder followed by a lightweight MLP decoder (See implementation details in the supplementary).
Offline training is performed using the Adam optimizer, with learning rates of $1 \times 10^{-2}$ for the mapping network and $5 \times 10^{-3}$ for the latent vectors, and is run for 50 iterations.
All models were implemented in PyTorch and trained on an NVIDIA A100 GPU. Detailed inference time analysis is provided in the supplementary material.

\label{sec:results}
\begin{table*}[t]
\centering
\begin{tabular}{clcccc} 
\toprule
\multirow{2.5}{*}{\textbf{Category}} 
& \multicolumn{1}{c}{\multirow{2.5}{*}{\textbf{Method}}} 
& \multicolumn{2}{c}{\textbf{XCAT Phantom}} 
& \multicolumn{2}{c}{\textbf{In-house Data}} 
\\ 
\cmidrule(lr){3-4} \cmidrule(lr){5-6}
& & \textbf{Immediate} & \textbf{After 2 min} & \textbf{Immediate} & \textbf{After 2 min} \\ 
\midrule

\multirow{2}{*}{\texttt{Analytical}}
& NUFFT~\cite{NUFFT} 
& 7.80/0.252 
& 7.79/0.252 
& 10.89{\scriptsize{$\pm$0.53}}/0.364{\scriptsize{$\pm$0.022}} 
& 10.90{\scriptsize{$\pm$0.52}}/0.365{\scriptsize{$\pm$0.021}}
\\
& GRASP~\cite{GRASP} 
& 8.47/0.158 
& 8.47/0.158 
& 10.89{\scriptsize{$\pm$0.60}}/0.120{\scriptsize{$\pm$0.017}}
& 11.05{\scriptsize{$\pm$0.88}}/0.126{\scriptsize{$\pm$0.027}}
\\
\midrule

\multirow{2}{*}{\shortstack{\texttt{Retrospective}\\\texttt{Recon.}}}
& TDDIP~\cite{yoo2021time} 
& 17.73/0.498 
& 18.05/0.552 
& 25.38{\scriptsize{$\pm$1.84}}/0.661{\scriptsize{$\pm$0.044}}
& 25.70{\scriptsize{$\pm$2.81}}/0.687{\scriptsize{$\pm$0.068}}
\\
& SPINER~\cite{chen2025single} 
& 20.25/0.873 
& 20.10/0.869 
& 35.43{\scriptsize{$\pm$1.89}}/0.942{\scriptsize{$\pm$0.018}}
& 36.36{\scriptsize{$\pm$3.40}}/0.946{\scriptsize{$\pm$0.022}}
\\
\midrule

\multirow{3}{*}{\shortstack{\texttt{Prospective}\\\texttt{Recon.}}}
& Prior-INR~\cite{liu2024volumetric} 
& 15.05/0.444 
& 15.27/0.473 
& 26.72{\scriptsize{$\pm$2.46}}/0.810{\scriptsize{$\pm$0.087}}
& 27.00{\scriptsize{$\pm$2.76}}/0.811{\scriptsize{$\pm$0.087}}
\\
& MR-MOTUS~\cite{MR-MOTUS}
& 24.39/0.931 
& 24.22/0.929 
& 41.04{\scriptsize{$\pm$3.41}}/0.981{\scriptsize{$\pm$0.007}}
& 41.11{\scriptsize{$\pm$3.50}}/0.976{\scriptsize{$\pm$0.011}}
\\
& \textbf{PDMR (Ours)} 
& \textbf{26.28}/\textbf{0.958}
& \textbf{25.52}/\textbf{0.950} 
& \textbf{46.32}{\scriptsize{$\pm$4.06}}/\textbf{0.994}{\scriptsize{$\pm$0.003}}
& \textbf{43.39}{\scriptsize{$\pm$4.58}}/\textbf{0.978}{\scriptsize{$\pm$0.023}}
\\
\bottomrule
\end{tabular}
\caption{Quantitative results (PSNR (dB)/SSIM) of compared methods on the XCAT phantom and in-house datasets under immediate and 2-minute delayed prospective reconstruction settings. The best results are highlighted in \textbf{bold}.}
\label{table:pro_result}
\end{table*}

\subsection{Prospective Reconstruction}
\paragraph{Experimental Settings.}
To evaluate the effectiveness of PDMR for prospective reconstruction, we conduct experiments on both XCAT phantoms and in-house abdominal MRI data under two prospective settings:
1) \textbf{Immediate}: the prospective acquisition starts immediately after the retrospective scan.
2) \textbf{After 2 min}: where a two-minute interval separates the retrospective and prospective acquisitions.  

In all experiments, the analytical algorithms are directly evaluated under the prospective reconstruction setting. Both the prospective methods and the retrospective baselines are first trained on the retrospective data, and then adapted to the newly acquired measurements during the prospective stage. We compare the dynamic MR reconstruction performance of all baselines in the prospective scenario. Additional implementation details, parameter settings, and evaluation protocols are provided in the supplement.

\paragraph{Comparisons of Reconstructed MR images.}
Figure~\ref{fig:pros} shows the quantitative results.
The analytical method GRASP results in highly blurred images, losing all anatomical details, highlighting the failure of traditional methods with ultra-sparse measurements in the latency window.
For retrospective methods, SPINER~\cite{chen2025single} fails to generalize to unseen time points. It tends to extrapolate past motion trends and eventually produces nearly static outputs when evaluated at new timestamps, indicating its limited ability for prospective reconstruction.
For the prospective methods, Prior-INR~\cite{liu2024volumetric}, which relies on a hand-crafted and discrete manifold, also exhibits noticeable limitations. As shown in the $z\!-\!t$ profile, the learned motion trajectories are discontinuous and fail to reflect the smooth and continuous nature of physiological motion, leading to inaccurate deformation estimation during prospective reconstruction.
MR-MOTUS shows better motion tracking, but still exhibits noticeable errors, which are particularly evident in the $z\!-\!t$ profile error comparison.
Additionally, due to its linear representation limitations, MR-MOTUS struggles to capture small motions, as highlighted with red arrows in the figure.
In contrast, PDMR achieves near-perfect alignment with the ground truth, accurately capturing both large-scale motion and fine local dynamics, demonstrating its superior ability to model complex deformation patterns with high fidelity.

\par Table~\ref{table:pro_result} presents the quantitative results.
PDMR outperforms all baselines in both settings across XCAT Phantom and in-house data, showing a significant improvement over the SOTA MR-MOTUS (\textit{e.g.}, $\sim$ 2dB gain on in-house data).
These results demonstrate that PDMR consistently provides more accurate and reliable reconstructions, highlighting its superior performance in prospective reconstructions across diverse acquisition conditions.

\subsection{Evaluation of PDMR's Effectiveness}

We evaluated PDMR's performance in both offline learning from retrospective data and prospective adaptation of DVFs.
\paragraph{Evaluation of Offline Learning.}
Table~\ref{table:retro_results} shows the performance of PDMR on retrospective data compared with baselines.  It outperforms advanced retrospective methods, demonstrating its ability to effectively learn motion dynamics and capture detailed deformation fields from undersampled data.
\paragraph{Evaluation of DVFs.}
Figure~\ref{fig:dvf} shows the estimated DVFs alongside the dynamic reconstruction results in the prospective setting. The quiver plots (second column) visualize the direction and magnitude of the predicted DVFs, which closely follow the expected organ motion patterns, demonstrating that our method accurately captures respiratory dynamics. 
Notably, for regions that remain stationary (such as the spine in the XCAT phantom) our method preserves the absence of motion, demonstrating that the learned deformation manifold does not introduce spurious displacements in anatomically static structures. Additional visual results are provided in the supplementary material.

\begin{table}[!t]
\centering
\setlength{\tabcolsep}{3.5pt}
\begin{tabular}{ccc}
\toprule
\textbf{Method} & \textbf{XCAT} & \textbf{In-house} \\
\midrule

\text{TDDIP}~\cite{yoo2021time}
& 17.21/0.372
& 25.93{\scriptsize{$\pm$2.07}}/0.582{\scriptsize{$\pm$0.058}}
\\

\text{SPINER}~\cite{chen2025single}
& 21.26/0.888
& 38.14{\scriptsize{$\pm$4.40}}/0.954{\scriptsize{$\pm$0.031}}
\\

\text{MR-MOTUS}~\cite{MR-MOTUS}
& 25.72/0.948
& 43.50{\scriptsize{$\pm$4.81}}/0.986{\scriptsize{$\pm$0.010}}
\\

\textbf{PDMR (Ours)}
& \textbf{26.63}/\textbf{0.959}
& \textbf{47.60}{\scriptsize{${\pm}$3.44}}/\textbf{0.995}{\scriptsize{${\pm}$0.002}}
\\

\bottomrule
\end{tabular}
\caption{
Quantitative results (PSNR (dB)/SSIM) of compared methods on the XCAT phantom and in-house datasets during retrospective learning. The best results are highlighted in \textbf{bold}.}
\label{table:retro_results}
\end{table}

\section{Discussion}
In this section, we analyze the interpretability and representativeness of the learned latent vector in PDMR (\S~\ref{sec:analysis_latent}). Meanwhile, we analyze the PDMR's ability to generalize to unseen motion in prospective reconstruction (\S~\ref{sec:unseen_motion}).

\begin{figure*}[t]
  \centering
   \includegraphics[width=\linewidth]{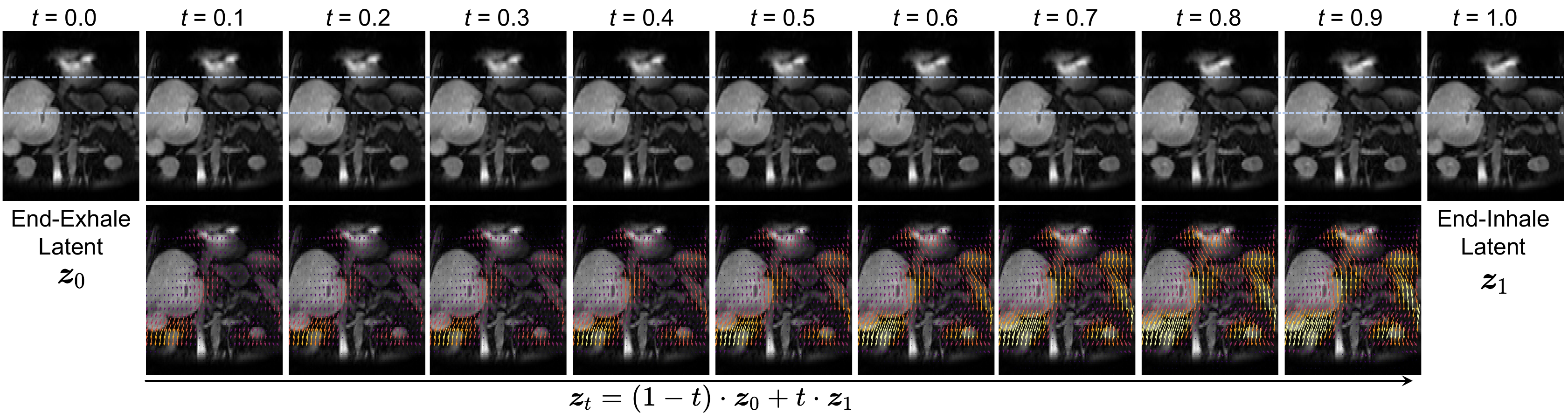}

   \caption{Visualization of reconstructed MR images and corresponding DVFs obtained from interpolating between two latent vectors, where $\boldsymbol{z}_0$ and $\boldsymbol{z}_1$ denote the learned vectors at the inhale and exhale motion states. The dashed line highlights the respiratory-induced displacement for clearer visualization. }
   \label{fig:interpolation}
\end{figure*}

\begin{figure}[t]
  \centering
   \includegraphics[width=\linewidth]{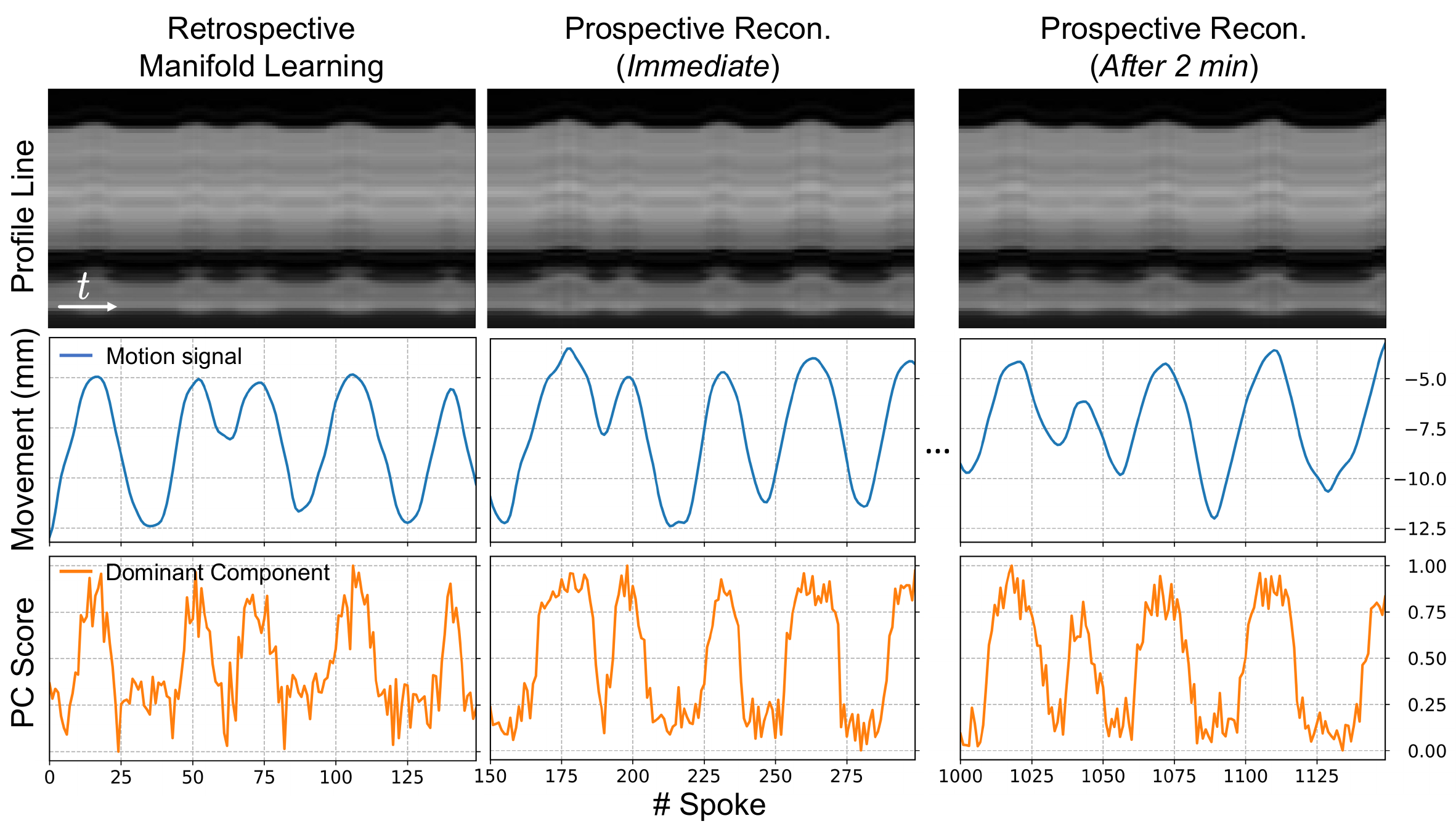}
   \caption{Visualization comparing the representative profile line ($z$–$t$ plane), the reference diaphragm motion, and the first principal component (PC) of the latent vector $\boldsymbol{z}$ in both retrospective and prospective reconstructions.}
   \label{fig:pca_latent}
\end{figure}

\subsection{Analysis of Latent Vector}
\label{sec:analysis_latent}

To demonstrate the interpretability of the latent vector, we perform PCA on the latent vectors $\boldsymbol{z}$ learned offline and adapted during prospective reconstruction. 
We extract the first principal component (PC) and compare it with the reference respiratory motion signal, which represents the superior–inferior displacement of the liver, shown in Figure ~\ref{fig:pca_latent}.
Despite small perturbations, the first PC of the latent code exhibits a clear correlation with the motion signal which indicates that the latent code learned by PDMR captures motion-related variations effectively.

To demonstrate that the learned manifold is continuous and physiologically reasonable, as well as to assess the generalization capability of the mapping network, we interpolate between latent vectors corresponding to the inhale and exhale respiratory states and decode them using the trained network.
Specifically, we denote the latent vector at the end-inhale state as $\boldsymbol{z}_0$ and at the end-exhale state as $\boldsymbol{z}_1$.
Intermediate latent vectors are generated through linear interpolation at uniform steps of 0.1 ($t = 0.1, \dots, 0.9$),
with the reconstructed images shown in Figure~\ref{fig:interpolation} (top row).
The highlighted horizontal lines visualize the inferred respiratory displacement, which varies smoothly with the interpolated latent vectors. The corresponding template images and DVFs are displayed in the bottom row of Figure~\ref{fig:interpolation}, and the consistent linear progression in the DVFs further indicates that the latent space encodes a continuous, structured representation of respiratory motion. Additional analysis is provided in the supplementary material.

\subsection{Robustness on Unseen Motion State}
\label{sec:unseen_motion}
To explore the performance of PDMR under unseen motion patterns, we introduce additional respiratory displacements (0–3 mm in the vertical direction) during prospective reconstruction and the quantitative result is shown in Table~\ref{table:unseen}.
The reconstruction quality gradually decreases as the displacement increases; however, even with a 3 mm offset, which is well outside the range observed during offline training, PDMR still achieves 38.80 dB PSNR, demonstrating strong robustness to unseen motion.

\begin{table}[t]
\centering
\begin{tabular}{ccccccc}
\toprule
 \textbf{Offset}& 0 mm & 1 mm & 2 mm & 3 mm \\
\midrule
\textbf{PSNR (dB)} & 52.53 & 44.65 & 40.63 & 38.80 \\
\textbf{SSIM} & 0.998    & 0.991 & 0.978 & 0.964\\
\bottomrule
\end{tabular}
\caption{Qualitative results of PDMR with unseen motion patterns with different additional motion offsets.}
\label{table:unseen}
\end{table}

\section{Conclusion}
\label{sec:conclu}
In this work, we introduced PDMR, a manifold-based framework for prospective dynamic 3D MRI reconstruction from single measurements. PDMR retrospectively learns a compact and temporally generalizable motion manifold with a geometry-aware tri-plane mapping network, enabling high-fidelity deformation modeling. During online reconstruction, it efficiently adapts to new motion states by optimizing only a low-dimensional latent code, achieving accurate DVF estimation within a few iterations. Experiments on XCAT phantoms and in-house datasets show that PDMR outperforms SOTA methods, providing higher reconstruction fidelity, smoother motion trajectories, and stronger robustness to unseen motion. These results demonstrate the potential of manifold-based deformation modeling for real-time, motion-aware MRI guidance.

\newpage

\paragraph{Acknowledgment.} 
This work was partially supported by NIH R01EB032825 and Siemens Healthineers NIH/NHLBI R01HL163030.
LS acknowledges funding support by National Science Foundation via grant IIS-2435746, Defense Advanced Research Projects Agency under contract No. HR00112520042.
We thank Xuanyu Tian for insightful discussions and valuable assistance in the preparation of this manuscript.

{
    \small
    \bibliographystyle{ieeenat_fullname}
    \bibliography{main}
}


\end{document}